\documentclass[manuscript]{acmart}
\AtBeginDocument{%
  }

\usepackage{amsmath}
\usepackage[linesnumbered,ruled]{algorithm2e}
\usepackage{subcaption}
\usepackage{booktabs}
\usepackage{multirow}
\usepackage{makecell}
\usepackage{enumitem}
\usepackage{tabularx}
\setlist[enumerate]{topsep=2pt,itemsep=2pt,parsep=0pt,partopsep=0pt}
\begin{document}

\title{Query- and Candidate-Aware Long User Sequence Modeling for Sponsored
Product Ranking in Search}

\author{Dongyue Xie, Sergey Kolbin, Manavender Malgireddy, Iurii Shcherbak, Weixing Tang, Arushi Jain,  Wanchen Shao,  Kurt Zimmer, Raja Hafiz Affandi, Matt Lindsay}
\email{{dxie, skolbin, mmalgireddy, ishcherbak, wtang, ajain31,wshao, kzimmer, raffandi, mlindsay}@wayfair.com}
\affiliation{%
  \institution{Wayfair LLC}
  \country{USA}
}

\renewcommand{\shortauthors}{Xie et al.}

\begin{abstract}
User sequence modeling is widely used in recommendation and advertising. Search ranking presents an additional challenge: the current query expresses immediate intent, while a candidate product determines which parts of a user's long-term history are relevant. We present a production sequence framework for sponsored-product ranking that augments recent behavior with two retrieval channels over long-term history: candidate-nearest-neighbor retrieval and query-nearest-neighbor retrieval. The selected actions are encoded together with dedicated query and candidate context tokens, while candidate-aware early fusion injects the candidate into every history token. Offline ablations show that dual retrieval improves major ranking metrics relative to the production baseline. In an online A/B test on sponsored-product search traffic, the model increased CTR, attributed conversion rate, and orders per impression.
\end{abstract}


\keywords{Sponsored Products, Advertising, Ranking Optimization, Transformer, User Sequence Modeling, Sequence-aware Ranking}


\maketitle

\section{Introduction}


User behavior sequences are a central source of personalization in industrial recommendation and advertising systems.
 Recent customer interactions have been shown to materially improve
click-through-rate (CTR) and conversion-rate (CVR) predictions \cite{sasrec,bert4rec,bst,transact,transactv2,tokenmixer_large,sort}.  However, recent actions capture only
a narrow window of intent. A customer’s longer history may contain semantically relevant products and durable
preferences that are absent from the latest session.

Using long-term behavior in production is difficult for two reasons. First, a raw history can contain tens of thousands of heterogeneous actions and cannot be processed exhaustively within an auction-time latency budget. Second, most historical actions are irrelevant to a particular scoring decision. Long-sequence models therefore commonly retrieve or compress a small target-aware subset before applying a more expressive interaction model \cite{pi2020sim,chen2021eta,chang2023twin,si2024twinv2,transactv2}.

Sponsored-product ranking on a search-results page has a distinct contextual structure. The current query provides an explicit, request-level expression of intent. Candidate-only retrieval may recover prior interactions similar to the product being scored, but it can miss actions that are highly relevant to the query and only indirectly related to that candidate. An effective search sequence module should therefore condition on both contexts while remaining efficient enough to score many candidates per request.

We present a production framework that combines recent behavior with candidate- and query-based retrieval from a long-term action pool. The selected actions are encoded with two dedicated context tokens: one for the current query and one for the candidate product. Our experiments isolate the effects of retrieval, global context tokens, early fusion, and an auxiliary next-action objective.
Our contributions are:
\begin{itemize}
    \item[(1)] We introduce dual contextual retrieval for sponsored-product search ranking, selecting long-term actions using both candidate-to-item and query-to-item semantic similarity.
    \item[(2)] We combine dedicated query and candidate tokens with candidate-aware early fusion.
    \item[(3)] We provide offline ablations and a production A/B test. The results show that candidate and query retrieval are complementary, early fusion is essential, and the online outcome improves engagement and conversion volume.
\end{itemize}
\section{Related Work}
 Transformer-based models have become standard for sequential recommendation: SASRec uses causal self-attention, BERT4Rec uses bidirectional masked-item training, and BST applies Transformer behavior modeling to e-commerce ranking \cite{sasrec,bert4rec,bst}. TransAct extends this line to real-time production ranking with heterogeneous user actions and candidate-aware token construction \cite{transact}. Recent industrial work also explores unified sequence and feature-interaction backbones at larger scale \cite{zhang2024wukong,zhang2026onetrans,tokenmixer_large,sort}.

Processing lifelong sequences requires selecting or compressing relevant behavior. TWIN aligns the relevance function used by coarse retrieval and exact target attention, while TWIN-V2 hierarchically compresses much longer histories \cite{chang2023twin,si2024twinv2}. TransAct V2 retrieves long-term actions and combines them with recent real-time behavior under production constraints \cite{transactv2}. LLaTTE studies scaling laws for long sequence models and moves expensive user modeling to an asynchronous upstream stage \cite{xiong2026llatte}. Our work differs in using two request-dependent retrieval views---the candidate product and the explicit search query---and in studying how these views interact with context tokens and token-level early fusion.

Direct production comparisons with TWIN, TWIN-V2, or TransAct V2 are
difficult because these systems were developed for different ranking stacks,
history representations, retrieval infrastructure, and serving constraints.
We therefore do not claim an apples-to-apples empirical comparison with these
methods. Instead, our experiments use the deployed recent-history model as
the production baseline and isolate the incremental effects of candidate-only
retrieval, query-only retrieval, and their combination within the same
ranking architecture and serving environment.

\section{Ranking Model}

In this section, we describe the production ranking model used at Wayfair and the sequence-augmented architecture studied in this work. 

\subsection{Production Ranking}
\label{sec:baseline_ranker}

\begin{figure}[t]
\centering
\includegraphics[width=0.5\columnwidth]{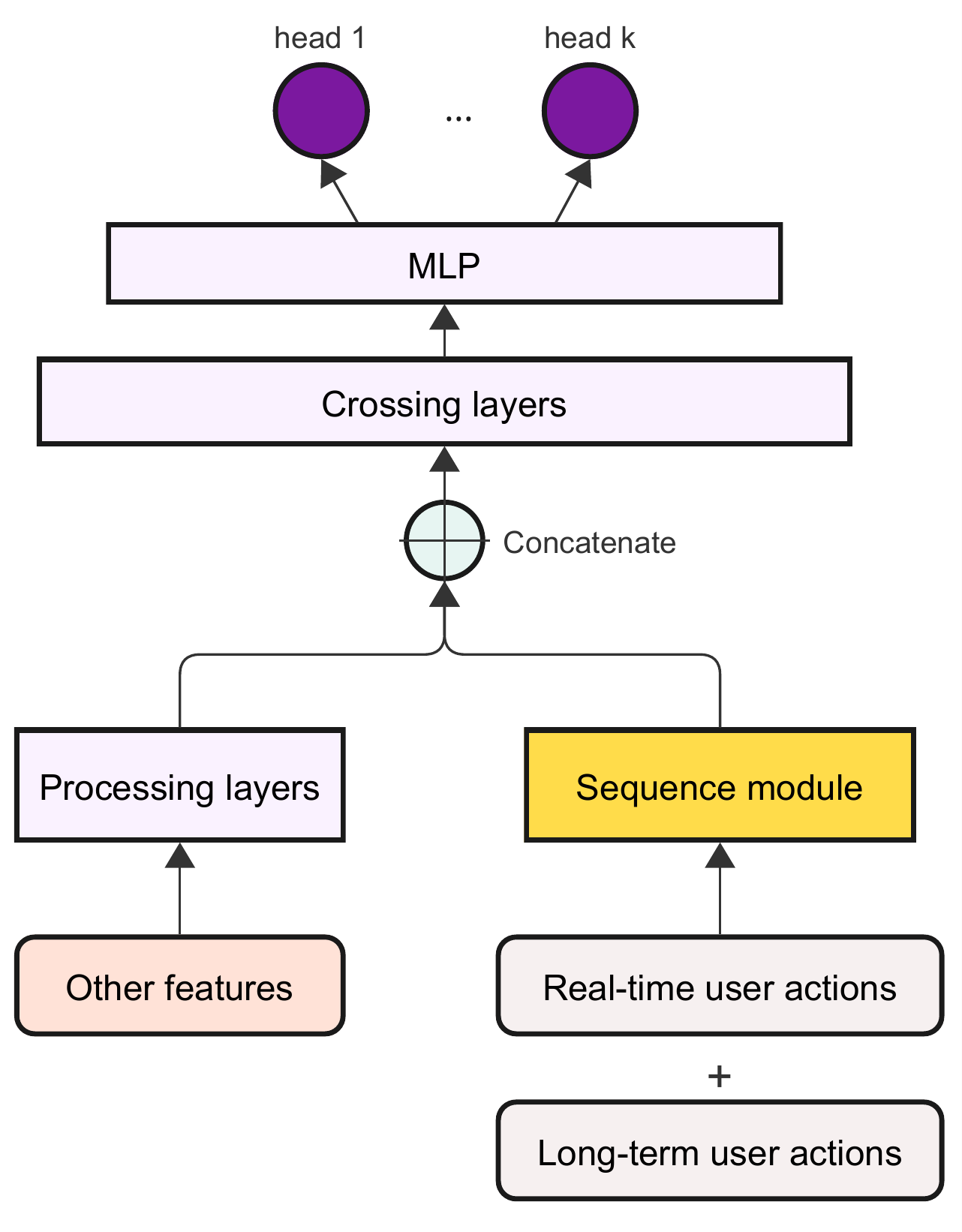}
\caption{Overview of the sponsored product ranking model. The proposed sequence module combines real-time actions with retrieved long-term actions.}
\label{fig:wayfair_ranking_overview}
\end{figure}

Wayfair's sponsored-products ranker is a multi-task deep neural network that estimates customer engagement and downstream conversion outcomes for each candidate product. The task heads include binary click and conversion targets and numerical outcomes. Predicted CTR and CVR are post-hoc calibrated with isotonic regression before being consumed by the downstream ranking and auction stack \cite{malgireddy2025profit}. The production baseline also includes a TransAct-style sequence module over recent heterogeneous user actions. Figure~\ref{fig:wayfair_ranking_overview} shows the high-level architecture.

\subsection{Sequence construction and tokenization}

For each ranking request, let $r$ denote the search query, $c$ the candidate SKU, and $\mathcal{H}_{\tau}$ the user's historical actions before request time $\tau$. Each historical action is represented as
$(a_j, s_j, t_j, p_j),$
where $a_j$ is the action type, $s_j$ is the interacted SKU, $t_j$ is the timestamp, and $p_j$ is the interaction surface.

We construct the model sequence from three complementary views of the user's history. Let \[ \mathcal{R} = \operatorname{Recent}_{L_{\mathrm{recent}}}(\mathcal{H}_{\tau}) \] denote the $L_{\mathrm{recent}}$ most recent actions. Candidate- and query-based retrieval are performed only over the remaining long-term history, \[ \mathcal{H}_{\tau}^{\mathrm{long}} = \mathcal{H}_{\tau} \setminus \mathcal{R}. \] From this retrieval pool, $\mathcal{C}$ contains the actions whose interacted SKUs are most similar to the candidate SKU $c$, and $\mathcal{Q}$ contains the actions whose interacted SKUs are most relevant to the current query $r$. The selected history is therefore \[ \mathcal{S} = \mathcal{R} \cup \mathcal{C} \cup \mathcal{Q}. \] An action selected by both retrieval sources is retained only once, using its unique event identifier. Consequently, the final history contains at most \[ L_{\mathrm{hist}} = L_{\mathrm{recent}} + L_{\mathrm{cand}} + L_{\mathrm{query}} \] historical action tokens, with fewer tokens possible when the retrieval views overlap.


The Transformer input contains two request-level context tokens followed by the selected historical action tokens:
\[
\mathbf{X}
=
\left[
\mathbf{x}_q;
\mathbf{x}_c;
\mathbf{x}_{1};
\ldots;
\mathbf{x}_{|\mathcal{S}|}
\right],
\]
where the query and candidate tokens are defined as
\[
\mathbf{x}_q =
LN\left(
\mathbf{q}
+
E_{\mathrm{role}}(\mathrm{query})
\right),
\]
\[
\mathbf{x}_c=
LN\left(
\mathbf{c}
+
E_{\mathrm{role}}(\mathrm{candidate})
\right),
\]
where $\mathbf{q}$ and $\mathbf{c}$ are the projected representations of the current query $r$ and candidate SKU $c$, respectively.

For each selected historical action $j$, the history token combines the action, historical SKU, candidate context, SKU metadata, interaction surface, and temporal information:
\[
\mathbf{x}_j=
LN\left(
\operatorname{MLP}_{\mathrm{token}}
\left[
E_a(a_j)
\Vert
\mathbf{u}_{j}
\Vert
\mathbf{c}
\Vert
E_{\mathrm{meta}}(\mathbf{u}_{j})
\Vert
E_{\mathrm{meta}}(\mathbf{c})
\Vert
E_{\mathrm{surf}}(p_j)
\right]
+
E_t(t_j)
+
E_{\mathrm{role}}(\mathrm{history})
\right),
\]
where $\mathbf{u}_j$ represents the historical SKU and $E_t(t_j)$ represents the recency and time gap of the action. Including $\mathbf{c}$ directly in every history token provides candidate-aware early fusion.

A Transformer encoder contextualizes the complete sequence:
\[
\mathbf{H}=
\operatorname{TransformerEncoder}(\mathbf{X}),
\]
with output
\[
\mathbf{H}=
\left[
\mathbf{h}_q;
\mathbf{h}_c;
\mathbf{h}_{1};
\ldots;
\mathbf{h}_{|\mathcal{S}|}
\right].
\]

\begin{figure}[t]
\centering
\includegraphics[width=0.45\columnwidth]{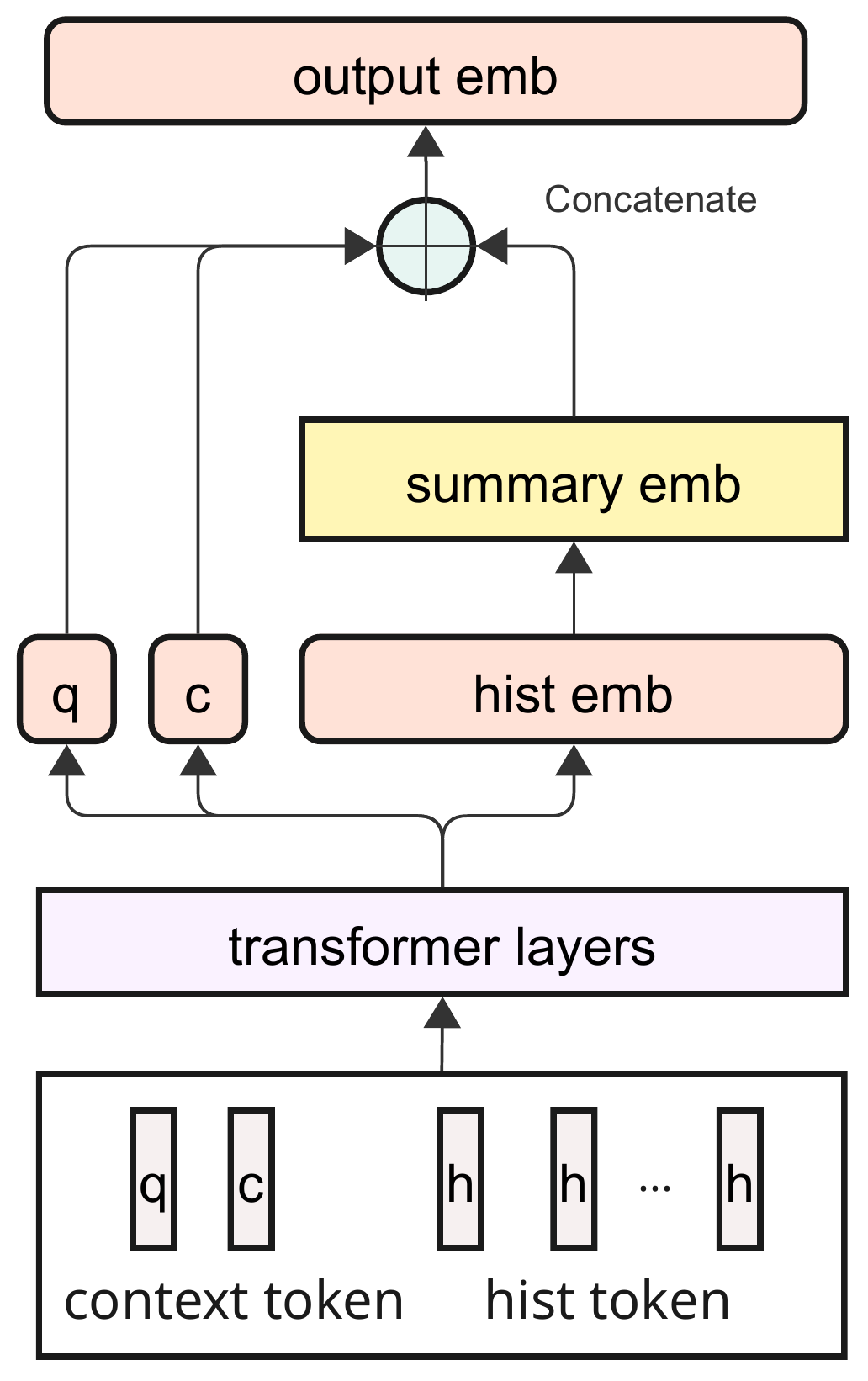}
\caption{Sequence module. Query and candidate context tokens are contextualized with retrieved history; the candidate is also fused
into every history token.}
\label{fig:wayseq_architecture}
\end{figure}

The Transformer outputs are summarized using learned sequence projection ($\mathbf{z}_{\mathrm{proj}}$), candidate-aware attention pooling ($\mathbf{z}_{\mathrm{candpool}}$), and masked max pooling ($\mathbf{z}_{\mathrm{max}}$) \cite{xie2026wayseq}:
\[
\mathbf{z}_{\mathrm{seq}}=
LN\left(
W_{\mathrm{seq}}
\left[
\mathbf{h}_q
\Vert
\mathbf{h}_c
\Vert
\mathbf{z}_{\mathrm{proj}}
\Vert
\mathbf{z}_{\mathrm{candpool}}
\Vert
\mathbf{z}_{\mathrm{max}}
\right]
\right).
\]
The resulting representation captures the current query and candidate together with both recent behavior and semantically relevant actions retrieved from the user's longer-term history. Figure~\ref{fig:wayseq_architecture} summarizes the architecture. 

\subsection{Training objectives}
\label{sec:loss}

We use a standard multi-task objective with binary cross-entropy for binary heads and mean-squared error for numerical heads. Let $\mathcal{H}_{\mathrm{bin}}$ and $\mathcal{H}_{\mathrm{num}}$ denote the binary and numerical task sets. The ranking loss is

\[
\mathcal{L}_{\mathrm{rank}}
=
\sum_{h\in\mathcal{H}_{\mathrm{bin}}}
\lambda_h
\left[
-\frac{1}{N}
\sum_{n=1}^{N}
\left(
y_{n,h}\log \hat y_{n,h}
+
(1-y_{n,h})\log(1-\hat y_{n,h})
\right)
\right]
+
\sum_{h\in\mathcal{H}_{\mathrm{num}}}
\lambda_h
\left[
\frac{1}{N}
\sum_{n=1}^{N}
(y_{n,h}-\hat y_{n,h})^2
\right].
\]

We additionally study an auxiliary next-action loss (NAL), following TransAct V2~\cite{transactv2}. NAL is applied only to the contiguous most recent-history sequence and is calculated using in-batch negatives (i.e. using the true next SKU as the positive and other next-action SKUs in the mini-batch as negatives). The complete objective is $\mathcal{L}_{\mathrm{rank}} + w_{\text{NAL}}\mathcal{L}_{\mathrm{NAL}}$.




A standard causal mask over history tokens is insufficient when query and candidate tokens attend globally. In a multi-layer Transformer, a context token could attend to a future target action in one layer and relay that information to an earlier history token in a later layer. We therefore use a hybrid attention design. In the first $M$ layers, history-to-history attention is causal; history tokens may attend to the query and candidate tokens; and the two context tokens may attend only to themselves and each other. NAL is computed from the history representations at layer $M$. In the remaining layers, attention is fully bidirectional, allowing query, candidate, and history tokens to form the final ranking representation. Only the ranking loss is applied to this global block.

\section{Offline Experiments}

We evaluate whether long-term retrieval, together with dedicated query and
candidate context tokens, improves contextual sequence ranking beyond a
recent-history-only model on a Wayfair internal dataset. We evaluate grouped
click AUC and order-7d AUC on a one-month held-out test dataset. Results are
reported relative to the baseline:
\begin{equation}
    \Delta(\%) =
    100 \times
    \frac{m - m_{\mathrm{Base}}}{m_{\mathrm{Base}}}.
\end{equation}
For AUC, a positive value is better. Because the calibration layer is shared across
variants and independent of the sequence module, the ablation
study focuses on discrimination metrics. Due to business confidentiality constraints, all results are presented as relative deltas rather than absolute values. Uncertainty estimates were not retained for these historical
offline runs. We therefore interpret small differences between closely
performing variants cautiously and focus on the larger and directionally
consistent effects across tasks. 

In both the offline experiments and final online serving system, the long-term user history contains \(O(10^3)\) actions within a one-year lookback window. From this history, we construct a sequence of up to 200 actions: the 100 most recent actions, 50 actions retrieved through candidate-nearest-neighbor search, and 50 actions retrieved through query-nearest-neighbor search. The 100-action recent-history budget is inherited from the existing production sequence model. We allocate 50 actions to each retrieval method to keep the total sequence length within the production serving budget. In our serving environment, a 200-action sequence remains practical. We therefore treat the 100/50/50 allocation as a production operating point rather than a globally optimal retrieval budget. The NAL loss weight is set to \(0.05\). Candidate-NN retrieval uses 32-dimensional SKU embeddings to limit ANN lookup
and memory cost in the online system. The query and SKU representations used for retrieval and sequence token
construction are produced by Wayfair's internal semantic-search models. Action-type, interaction-surface, and metadata embeddings are also 32-dimensional and are learned jointly with the ranking model. The sequence encoder contains $2$ Transformer layers with
hidden dimension $128$ and $2$ attention heads.
The first layer uses the hybrid causal attention pattern described in
Section \ref{sec:loss}, while the second layer uses fully bidirectional
attention. The feed-forward dimension is $64$ and dropout is set to
$0.1$. Table~\ref{tab:configs} lists the ablation variants.

\begin{table}[t]
\caption{Ablation configurations. The baseline is the production recent-history sequence model.}
\label{tab:configs}
\centering
\small
\begin{tabular}{lllll}
\toprule
ID & Retrieval history & $x_q/x_c$ & Early fusion & NAL \\
\midrule
baseline & Recent only & Off & On & Off \\
M0 & Recent only & On & On & On \\
M1 & Recent + candidate-NN & On & On & On \\
M2 & Recent + query-NN & On & On & On \\
M3 & Recent + candidate-NN + query-NN & On & On & On \\
M4 & Recent + candidate-NN + query-NN & Off & On & On \\
M5 & Recent + candidate-NN + query-NN & On & On & Off \\
M6 & Recent + candidate-NN + query-NN & On & Off & On \\
\bottomrule
\end{tabular}
\end{table}

\begin{table}[t]
\caption{Offline performance relative to the production baseline.}
\label{tab:offline}
\centering
\small
\begin{tabular}{lrr}
\toprule
ID & $\Delta$ Click AUC (\%) & $\Delta$ Order-7d AUC (\%) \\
\midrule
baseline & -- & -- \\
M0 & +0.46 & -0.61 \\
M1 & +1.16 & +1.35 \\
M2 & +1.00 & +1.16 \\
M3 & \textbf{+1.37} & \textbf{+2.19} \\
M4 & +0.45 & +0.28 \\
M5 & +1.11 & +1.46 \\
M6 & -4.71 & -6.06 \\
\bottomrule
\end{tabular}
\end{table}

\subsection{Results}

Table~\ref{tab:offline} shows that adding the context tokens and NAL
to recent history alone (M0) provides only a small click-AUC improvement
(+0.46\%) and slightly reduces order-7d AUC (-0.61\%). We therefore do not
interpret the context tokens or auxiliary objective as being independently
sufficient to improve all ranking tasks. Instead, their benefit is strongest
when combined with relevant long-term behavior.

The main improvement comes from retrieving relevant long-term actions.
Candidate-NN retrieval alone (M1) improves click AUC by 1.16\% and order-7d
AUC by 1.35\%, while query-NN retrieval alone (M2) improves them by 1.00\%
and 1.16\%, respectively. The dual-retrieval model (M3) achieves the largest
observed improvements, with gains of 1.37\% in click AUC and 2.19\% in
order-7d AUC. These results suggest that query- and candidate-based retrieval
provide complementary views of long-term user behavior.

Removing the dedicated query and candidate tokens (M4) reduces the click-AUC gain from 1.37\% to 0.45\%. The context tokens therefore add information beyond candidate-aware token construction. Removing NAL (M5) reduces the click-AUC gain to 1.11\% and the order-7d AUC
gain to 1.46\%, suggesting a smaller positive contribution from the auxiliary
objective.

The largest effect comes from early fusion. Disabling candidate-aware early
fusion (M6) decreases click AUC by 4.71\% and order-7d AUC by 6.06\%
relative to the baseline. Importantly, early fusion is already present in the
production baseline and M6 removes an existing candidate-history
interaction mechanism rather than simply removing an additional component
introduced by the proposed model. Without
token-level candidate conditioning, the model must recover candidate-history
compatibility indirectly through the global candidate token and subsequent
self-attention. The large degradation suggests that this indirect interaction
is insufficient for the production ranking task. We additionally inspected the training and validation curves of the ablation
variants. M0 and M6 showed normal optimization behavior without evidence of
training divergence or premature termination. We therefore attribute their
performance differences primarily to the architectural changes rather than to
failed optimization.

\section{Online Experiments}

We evaluated the model in a five-week online A/B test on live sponsored-product search traffic. The experiment used user-level randomization and compared the model variant with the production control. The model was deployed within the latency budget of the existing sponsored
product ranking system. Among the newly introduced components, candidate-NN fetching and longer sequence are major contributors to the latency increase, although overall serving latency is still dominated by fetching upstream model outputs used as ranking features. End-to-end model
serving latency increases by approximately 12\% relative to the production
baseline.


Table~\ref{tab:online} reports relative treatment effects from the production experimentation platform. Absolute traffic volumes and metric values are omitted due to business confidentiality. Statistical significance was assessed by the production experimentation platform using a significance threshold of $\alpha=0.05$.

\begin{table}[t] 
\centering 
\caption{Relative change of the proposed model versus the production control in online sponsored-product search traffic. Statistical significance is evaluated at $\alpha=0.05$.} 
\label{tab:online}
\begin{tabular}{lcc} \toprule Metric & Relative change (\%) & Significant at $\alpha=0.05$ \\ \midrule CTR & +6.0 & Yes \\ CPC & -1.3 & Yes \\ Ad spend & +0.8 & Yes \\ Orders / impression & +11.5 & Yes \\ Attributed CVR & +5.2 & Yes \\ \bottomrule \end{tabular} \end{table}


The treatment increased CTR, attributed CVR, and orders per impression, while CPC decreased and total ad spend increased modestly. These outcomes are consistent with improved matching and higher conversion volume.

\section{Discussion and Future Work}
Our experiments provide three main findings. First, retrieving semantically relevant actions from long-term history adds substantial value beyond recent behavior. Second, candidate-NN shows slightly larger observed gains when used alone, while adding
query-NN retrieval yields the largest observed gains in the dual-retrieval
configuration. Third, global query and candidate tokens do not make token-level candidate conditioning redundant.  While the retrieval budgets, embedding dimensions, and sequence length reflect
a production operating point rather than an exhaustive hyperparameter search,
and our empirical comparisons focus on controlled ablations within Wayfair's
production architecture rather than independent reimplementations of other
industrial lifelong-sequence systems, these choices allow us to isolate the
contribution of query- and candidate-aware retrieval under a consistent
serving environment. Accordingly, our conclusions are limited with respect to
globally optimal architectures and hyperparameter choices.

Several directions follow from these results. First, we plan to study scaling along history-pool size, retrieved-sequence length, embedding dimension, and Transformer depth while preserving serving latency. Efficient attention, cached projections, hierarchical compression, and asynchronous user encoders are promising mechanisms for this regime \cite{si2024twinv2,xiong2026llatte}. Second, the quality of retrieval depends on the 
representations used to match the current query and candidate product to historical products. 
In the current system, these representations are produced by existing
semantic-search models and are fixed with respect to the ranking model.
Improving these representations could therefore further improve the quality
of retrieved user behavior. In particular, query and product representations
could be strengthened through multimodal or foundation-model pretraining and
then distilled into compact encoders. Third, retrieval budgets can be made adaptive to query specificity, candidate novelty, history density, action type, and recency. Multi-interest retrieval may also improve coverage when a customer's lifelong history contains several distinct intents.

\bibliographystyle{ACM-Reference-Format}
\bibliography{wayseq}

\end{document}